\documentclass[letterpaper]{article} % DO NOT CHANGE THIS
\usepackage{aaai2026}  % DO NOT CHANGE THIS
\usepackage{times}
\usepackage{helvet}
\usepackage{courier}
\usepackage[hyphens]{url}
\usepackage{graphicx}
\usepackage{booktabs}
\usepackage{multirow}
\usepackage[table]{xcolor}
\usepackage{arydshln}
\usepackage{pifont}

\usepackage{adjustbox}
\usepackage{amsmath}
\usepackage{amssymb}
\usepackage{bm}
\definecolor{c1}{HTML}{003371}
\usepackage{natbib}
\usepackage{caption}
\usepackage{cleveref}
\usepackage{makecell}
\usepackage{cellspace}
\usepackage[table]{xcolor}
\usepackage{booktabs}
\usepackage[algo2e]{algorithm2e}
\usepackage[T1]{fontenc}
\usepackage{algpseudocode}
\usepackage{amsmath}
\usepackage[most]{tcolorbox}
\definecolor{blue1}{HTML}{013371}
\newtcbtheorem[]{trainprompt}{Case}%
{colback=gray!00,colframe=blue1,fonttitle=\bfseries, left=.02in, right=.02in,bottom=.02in, top=.02in}{trainprompt}
\usepackage{algorithm}
\usepackage{tabularx}

\usepackage{newfloat}
\usepackage{listings}
\DeclareCaptionStyle{ruled}{labelfont=normalfont,labelsep=colon,strut=off}
\floatstyle{ruled}
\newfloat{listing}{tb}{lst}{}
\floatname{listing}{Listing}
\title{CausalStep: A Benchmark for Explicit Stepwise Causal Reasoning in Videos}
\author{
    Xuchen Li\textsuperscript{\rm 1,}\textsuperscript{\rm 2,}\textsuperscript{\rm 3}\thanks{Equal contribution.},
    Xuzhao Li\footnotemark[1],
    Shiyu Hu\textsuperscript{\rm 4},
    Kaiqi Huang\textsuperscript{\rm 1,2}\thanks{Correspondence authors.},
    Wentao Zhang\textsuperscript{\rm 3,5}\footnotemark[2]
}
\affiliations{
    \textsuperscript{\rm 1}CASIA,
    \textsuperscript{\rm 2}UCAS,
    \textsuperscript{\rm 3}ZGCA,
    \textsuperscript{\rm 4}NTU,
    \textsuperscript{\rm 5}PKU,

    s-lxc24@bjzgca.edu.cn, kqhuang@ia.ac.cn, wentao.zhang@pku.edu.cn
}

\usepackage{bibentry}
\begin{document}

\maketitle

\begin{abstract}
Recent advances in large language models (LLMs) have improved reasoning in text and image domains, yet achieving robust video reasoning remains a significant challenge. Existing video benchmarks mainly assess shallow understanding and reasoning and allow models to exploit global context, failing to rigorously evaluate true causal and stepwise reasoning. We present CausalStep, a benchmark designed for explicit stepwise causal reasoning in videos. CausalStep segments videos into causally linked units and enforces a strict stepwise question-answer (QA) protocol, requiring sequential answers and preventing shortcut solutions. Each question includes carefully constructed distractors based on error type taxonomy to ensure diagnostic value. The benchmark features 100 videos across six categories and 1,852 multiple-choice QA pairs. We introduce seven diagnostic metrics for comprehensive evaluation, enabling precise diagnosis of causal reasoning capabilities. Experiments with leading proprietary and open-source models, as well as human baselines, reveal a significant gap between current models and human-level stepwise reasoning. CausalStep provides a rigorous benchmark to drive progress in robust and interpretable video reasoning.
\end{abstract}

\section{Introduction}
\label{sec:intro}
\begin{table*}[htbp!]
  \centering
  \caption{Comparison between CausalStep and existing video understanding/reasoning benchmarks across key aspects: the number of videos (\textbf{\#Videos}), video duration (\textbf{Duration}), number of reasoning QA pairs (\textbf{\#QA Pairs}), spatio-temporal relationship understanding (\textbf{Spatio-temporal}), causal relationship understanding (\textbf{Causal}), stepwise reasoning protocol (\textbf{Stepwise}), and annotation methodology (\textbf{Annotation}). \textbf{A} denotes AI-generated, \textbf{M} denotes manual, and \textbf{A\&M} indicates a combination.}
  \begin{tabularx}{\textwidth}{Xccccccc}
    \toprule
    \textbf{Benchmark} & \textbf{\#Videos} & \textbf{Duration} & \textbf{\#QA Pairs} & \textbf{Spatio-temporal} & \textbf{Causal}  & \textbf{Stepwise} & \textbf{Annotation} \\
    \midrule
    MVBench & 200 & 15-20 s & 4,000 & \checkmark & \checkmark & \text{\sffamily X} & A \\
    TempCompass & 410 & 15-20 s & 7,540 & \checkmark & \text{\sffamily X} & \text{\sffamily X} & A\&M \\
    Video-MMMU & 300 & 506.2 s & 900 & \text{\sffamily X} & \text{\sffamily X} & \text{\sffamily X} & M \\
    MMVU & 1,529 & 51.4 s & 3,000 & \text{\sffamily X} & \text{\sffamily X} & \text{\sffamily X} & M \\
    Video-MME & 900 & 35.7 s & 1,944 & \checkmark & \checkmark & \text{\sffamily X} & M \\
    VCR-Bench & 859 & 159 s & 1,034 & \checkmark & \checkmark & \text{\sffamily X} & A\&M \\
    Video-Holmes & 270 & 160 s & 1,837 & \checkmark & \checkmark & \text{\sffamily X} & A\&M \\
    MMR-V & 317 & 277 s & 1,257 & \checkmark & \checkmark & \text{\sffamily X} & A\&M \\
    \midrule
    \textbf{Ours} & \textbf{100} & \textbf{430.5 s} & \textbf{1,852} & \checkmark & \checkmark & \checkmark & A\&M \\
    \bottomrule
  \end{tabularx}
  \label{tab:benchmark_cmp}
  \vspace{-10pt}
\end{table*}

Recent advances in large language models (LLMs) have driven impressive progress in text \cite{tusurvey}, image \cite{iu}, and general video understanding \cite{vusurvey}. However, extending these reasoning capabilities to complex, real-world video scenarios \cite{vrsurvey} remains a major challenge. Video reasoning is fundamentally different from text or static images, as videos encode rich, sequential, and multimodal information that requires models to perform long-range, multi-frame reasoning and evidence integration across both temporal and spatial dimensions. This capability is essential for applications such as embodied intelligence \cite{eisurvey}, intelligent surveillance \cite{issurvey}, and human-computer interaction \cite{hcibook}.

Despite recent progress, existing video reasoning benchmarks \cite{mvbench,tempcompass,videommmu,mmvu,mmrv,videoholmes} exhibit key limitations. Most benchmarks focus on perception or shallow understanding, requiring only the identification of relevant frames or context. Crucially, by typically providing the entire video as input, these benchmarks allow models to exploit global information or shortcut strategies, thereby failing to assess true causal and stepwise reasoning. As a result, they do not capture the causally grounded reasoning processes humans naturally employ when interpreting complex video narratives. Moreover, the design of distractor options in multiple-choice questions is often unsystematic, lacking systematic coverage of common reasoning errors and thus failing to rigorously challenge model robustness.

To address these gaps, we introduce \textbf{CausalStep}, a new benchmark specifically designed to evaluate \textit{explicit stepwise causal reasoning} in videos. In CausalStep, each video is manually segmented into a sequence of causally linked segments. At each step, the model is given the current and previous segments (if any), without access to future information, and must answer a question—either a descriptive understanding question or an explicit causal reasoning question—before it can access the next. This protocol strictly enforces sequential, causally dependent reasoning and precludes the use of global shortcuts. Furthermore, we design a novel distractor generation strategy: for each multiple-choice question, distractor options are systematically constructed according to a taxonomy of error types, including temporal confusion, causal misattribution, and object misrecognition. This ensures each question not only tests surface-level perception but also challenges the model’s ability to distinguish between plausible but incorrect alternatives.

CausalStep comprises 100 videos spanning six diverse categories (e.g., cartoons, movies, sports, performances, documentaries, and TV shows), totaling 1,852 multiple-choice question-answer (QA) pairs. Each question is carefully annotated and reviewed, covering both descriptive understanding and explicit stepwise causal reasoning tasks—enabling fine-grained analysis of models’ causal reasoning abilities. To provide a comprehensive assessment of model performance, we propose a suite of seven diagnostic metrics: chain success rate, average and maximum chain length, restart frequency, weighted score, and dedicated accuracies for descriptive understanding and isolated causal reasoning. These metrics capture not only overall accuracy but also the depth, stability, and robustness of a model’s reasoning process.

We conduct extensive experiments on CausalStep, evaluating a wide range of state-of-the-art proprietary and open-source multimodal models—including the latest GPT \cite{gpt41,gpt4o,gpt4omini}, Gemini \cite{gemini,gemini2.5}, Claude \cite{claude35}, Qwen \cite{qwen25}, Gemma \cite{gemma3}, InternVL \cite{internvl,internvl3}, LLaVA \cite{llavaov,llavavideo,videollava}, and Phi \cite{phi4} series—as well as human participants. Our results reveal a substantial gap between current models and human-level performance, especially in explicit stepwise causal reasoning. This disparity is primarily driven by models' difficulty in maintaining continuous, error-free reasoning chains and their vulnerability to subtle distractors. These results show that even the strongest models struggle with long-range causal integration and are susceptible to confusable distractors, highlighting the need for further advances in video reasoning of multimodal large language models (MLLMs).

Our main contributions are as follows:

\begin{itemize}
\item \textbf{A novel benchmark for explicit stepwise causal reasoning in videos:} We introduce CausalStep, which segments videos into causally linked units and enforces a strict stepwise QA protocol, enabling rigorous evaluation of sequential, causally grounded reasoning in complex video narratives.

\item \textbf{A comprehensive annotation and evaluation framework:} We design a hybrid annotation pipeline combining LLM generation and human review, and propose a taxonomy-based distractor generation strategy. We further introduce seven diagnostic metrics that provide a fine-grained, multi-dimensional assessment of model performance, covering reasoning depth, stability and robustness.

\item \textbf{Extensive empirical analysis and insights:} We benchmark a diverse set of state-of-the-art (SOTA) proprietary and open-source models, as well as human baselines, on CausalStep. Our experiments reveal a significant gap between current models and human-level stepwise causal reasoning, and provide actionable insights for future research on robust and interpretable video reasoning systems.
\end{itemize}

\section{Related Work}
\label{sec:related}

\begin{figure*}[t!]
  \centering   
  \includegraphics[width=0.95\linewidth]{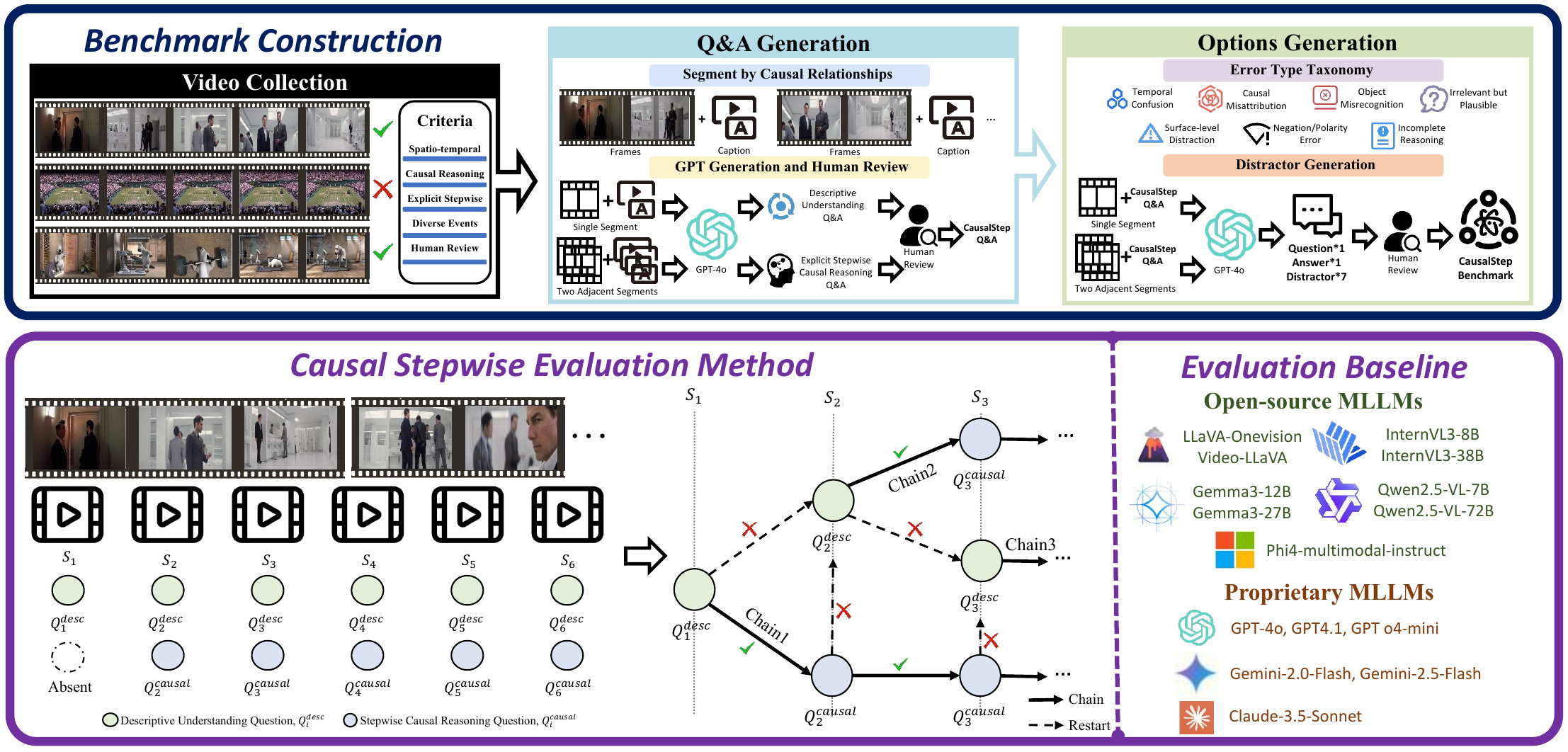}
  \caption{Overview of the CausalStep Benchmark Construction and Evaluation Framework. The top panel illustrates the benchmark construction pipeline: \textbf{Video Collection} with specific filtering, \textbf{Q\&A Generation} leveraging GPT-4o and human review for descriptive and causal questions, and \textbf{Options Generation} using a novel error type taxonomy. The bottom panel details the \textbf{Evaluation Method}, employing a strictly sequential, stepwise QA protocol. It also visualizes the \textbf{Causal Stepwise Evaluation} process with its enforced chain dependencies and restart mechanism, and lists the \textbf{Evaluation Baseline} models (open-source and proprietary MLLMs) used in our experiments.}
  \label{fig:overview}
  \vspace{-10pt}
\end{figure*}

\subsection{MLLMs for Video Understanding and Reasoning}
Recent progress in MLLMs has significantly advanced the field of video understanding and reasoning \cite{vusurvey,cao2025large,li2025verifybench}. Building on breakthroughs in image-based multimodal reasoning, models such as Gemma \cite{gemma3}, LLaVA-Onevision \cite{llavaov}, Phi \cite{phi4}, InternVL \cite{internvl}, Video-LLaVA \cite{llavavideo}, Qwen-VL \cite{qwen25}, GPT series \cite{gpt4o}, Claude \cite{claude35} and Gemini \cite{gemini} have adapted LLMs to process videos as sequences of frames, enabling temporal and contextual analysis. This typically involves utilizing visual encoders to extract frame features, which are then integrated with linguistic components. Despite these advancements, current MLLMs still face challenges in performing long-range, stepwise causal reasoning, especially when required to integrate information across multiple frames and modalities.

\subsection{Video Understanding Benchmark}
Video understanding benchmarks \cite{vusurvey,fiova,dtllm,dtvlt,vltverse} have evolved from early datasets on basic perception—like action recognition \cite{arsurvey,vltmi} and short-clip QA (e.g., MSRVTT-QA \cite{msrvttqa})—to broader evaluations. Recent benchmarks cover diverse content, tasks, and longer videos. Video-MME \cite{videomme} and MVBench \cite{mvbench} expand to multiple task formats and longer videos. LVBench \cite{lvbench} and LongVideoBench \cite{longvideobench} introduce long-form video QA, while MLVU \cite{mlvu} offers multi-task long-video understanding. E.T. Bench \cite{etbench} targets open-ended event-level reasoning, and TemporalBench \cite{temporalbench} evaluates fine-grained temporal analysis. InstructionBench \cite{instructionbench} tests step-by-step instructional videos for event/object-level reasoning. However, these benchmarks primarily evaluate perceptual and intuitive understanding, often not requiring complex, multi-step reasoning or integration of information across distant frames.

\subsection{Video Reasoning Benchmark}
The push toward more challenging video reasoning tasks \cite{vot} has led to the development of specialized benchmarks that go beyond perception. TempCompass \cite{tempcompass} and MVBench \cite{mvbench} focus on temporal and implicit causal reasoning in videos. In parallel, MMVU \cite{mmvu} and VideoMMMU \cite{videommmu} focus on reasoning in scientific or educational contexts, while VSI-Bench \cite{vsibench} targets indoor scene reasoning. Earlier work like VCR-Bench \cite{vcrbench} pioneered explicit chain-of-thought reasoning in videos, whereas STAR \cite{star} and VideoVista \cite{videovista} evaluate situated reasoning and versatile video QA, respectively. Adding depth, SOK-Bench \cite{sokbench} integrates open-world knowledge. Video-Holmes \cite{videoholmes} further decomposes spatio-temporal reasoning and multi-clue integration across a video. Nevertheless, there remains a gap in benchmarks that rigorously evaluate explicit, stepwise causal reasoning across long video sequences—a gap that CausalStep aims to address.

\section{CausalStep Task Overview}

The overview of CausalStep is illustrated in Figure~\ref{fig:overview}. In CausalStep, we propose an explicit stepwise causal reasoning task designed to rigorously evaluate a model’s ability to perform human-like, sequential causal reasoning over video content. The task is defined by the following components:

\textbf{Causal Segmentation of Videos.}
Each video is manually segmented into a sequence of causal segments, based on its underlying narrative and event structure in MGIT \cite{mgit}. A video is thus divided into $N$ segments, denoted as $\{S_1, S_2, \ldots, S_N\}$, where each segment corresponds to a distinct causal event or state. This segmentation is designed to support subsequent stepwise causal reasoning.

\textbf{Question Types.}
For every segment, we annotate a descriptive understanding question ($Q^{desc}_i$) that assesses the model’s comprehension of the observable content. These descriptive questions serve as the foundational starting point for each reasoning chain. For each segment except the first ($S_i$, $i>1$), we further annotate an explicit stepwise causal reasoning question ($Q^{causal}_i$), which requires the model to reason about the causal relationship between the current segment ($S_i$) and its direct preceding segment ($S_{i-1}$). Note that the first segment ($S_1$) does not have a causal reasoning question.

\textbf{Stepwise Reasoning Chain.}
The reasoning chain for evaluation begins with the descriptive understanding QA for the first segment ($S_1$). If the current descriptive understanding question ($Q^{desc}_i$) is answered correctly, the model will receive a score for that question, and the evaluation chain will immediately proceed to the explicit stepwise causal reasoning QA in the subsequent segment ($S_{i+1}$). If any answer is incorrect, the reasoning chain is interrupted (see Restart Mechanism). At each step with a causal reasoning QA ($Q^{causal}_i$), the model is provided with the current segment ($S_i$) and its direct preceding segment ($S_{i-1}$), along with the history of its previously correct answer. It cannot access future segments or questions in advance, strictly enforcing a stepwise progression and dependence on prior correct inferences.

\textbf{Restart Mechanism.}
If the model answers the current explicit stepwise causal reasoning QA incorrectly (e.g., at segment $S_k$), the current reasoning chain is interrupted, and it must restart from the descriptive understanding QA of the same segment ($S_k$). Whereas if the model answers any descriptive understanding QA incorrectly, the reasoning chain is interrupted, and it must restart from the descriptive understanding QA of the next segment ($S_{k+1}$) to initiate a new reasoning chain. This mechanism prevents accidental progression and ensures validity at each step of a successful causal chain.

\textbf{Scoring Scheme.}
Each correct descriptive understanding question is assigned a fixed score of 1 point. Each correct causal reasoning question's score is tied to its position within the current uninterrupted reasoning chain: the first causal question in a chain is worth 1 point, the second 2 points, and so on. If a reasoning chain is interrupted and restarted, the scoring for subsequent causal questions resets to 1 for the newly initiated chain. This system rewards models for maintaining longer correct reasoning sequences and provides a fine-grained measure of their explicit stepwise causal reasoning ability.

This task design establishes a rigorous protocol for evaluating models on explicit, sequential, and causally grounded reasoning, preventing shortcut solutions and mirroring human stepwise understanding. Algorithm \ref{algo:evaluation} details the full evaluation process, ensuring transparency and reproducibility.

\begin{algorithm}[t!]
\caption{CausalStep Evaluation Framework}
\label{algo:evaluation}
\KwIn{
  \\Segments $[S_1, S_2, \ldots, S_N]$;\\
  Descriptive QA list $[Q^{desc}_1, Q^{desc}_2, \ldots, Q^{desc}_N]$;\\
  Reasoning QA list $[Q^{causal}_2, \ldots, Q^{causal}_N]$;\\
  Model \texttt{M}
}
\KwOut{Total score for the video}

\SetKwFunction{Answer}{Answer} % M.Answer now returns (answer_content, is_correct_boolean)
\SetKwFunction{isCorrect}{is_correct} % Assuming is_correct is a helper function for clarity

$score \leftarrow 0$;

$chain\_length \leftarrow 0$; % Represents the current position/score multiplier in the chain

$i \leftarrow 1$;

$current\_question\_type \leftarrow \text{‘desc’}$;

\While{$i \leq N$}{
  \If{$current\_question\_type == \text{‘desc’}$}{
    $desc\_ans \leftarrow \text{\texttt{M.Answer}}(Q^{desc}_i, S_i)$\;

    \If{$is\_correct(desc\_ans)$}{
      $chain\_length \leftarrow chain\_length + 1$;
      
      $score \leftarrow score + 1$;

      $i \leftarrow i + 1$;

      $current\_question\_type \leftarrow \text{‘causal’}$;
    }\Else{
      $chain\_length \leftarrow 0$\tcp*{Restart}

      $i \leftarrow i + 1$;

      $current\_question\_type \leftarrow \text{‘desc’}$;
    }
  }

  \If{$current\_question\_type == \text{‘causal’}$}{
    \If{$i > N$}{ \textbf{break}; }

    $causal\_ans \leftarrow \text{\texttt{M.Answer}}(Q^{causal}_i, A_{i-1}, [S_{i-1}, S_i])$\;

    \If{$is\_correct(causal\_ans)$}{
      $chain\_length \leftarrow chain\_length + 1$;

      $score \leftarrow score + chain\_length$;

      $i \leftarrow i + 1$;

      $current\_question\_type \leftarrow \text{‘causal’}$;
    }\Else{
      $chain\_length \leftarrow 0$\tcp*{Restart}

      $i \leftarrow i + 1$;
      
      $current\_question\_type \leftarrow \text{‘desc’}$;
    }
  }
}
\textbf{return:} $score$;
\end{algorithm}

\section{CausalStep Benchmark Construction}
\label{sec:bench}

\subsection{Video Collection}
\subsubsection{Video Data Sourcing and Filtering.}
CausalStep draws inspiration from the recently proposed MGIT \cite{mgit} benchmark for its video data collection. MGIT references film narrative principles for video selection, focusing on causal event changes across temporal and spatial dimensions. It comprises 150 long video sequences with rich spatio-temporal and causal relationships, manually annotated at three semantic granularities (action, activity, and story). While MGIT itself is not designed for stepwise reasoning, its action-level segmentation specifically provides a foundation for constructing our explicit stepwise causal reasoning tasks.

Building upon MGIT, we curate a subset of videos specifically for stepwise causal reasoning evaluation. Our filtering prioritizes videos that: (1) support explicit stepwise causal reasoning with interconnected events across temporal segments; (2) discourage shortcut solutions where answers can be inferred from a single scene; and (3) feature key events distributed across different times and/or locations to prevent reliance on local context. This results in a more challenging foundation for CausalStep. We retain 100 videos, averaging 430.5 seconds each, spanning six diverse categories: Cartoons, Movies \& TV Shows, Outdoor Sports, Regular Sports, Performances, and Documentaries, ensuring broad coverage of real-world scenarios for explicit stepwise causal reasoning.

\subsection{Video Annotation}
We employ a hybrid annotation process, combining the efficiency of LLMs with the quality control of human review, to generate high-fidelity QA pairs. For the relevant prompts and human review principles, please refer to Appendix B and C.

\subsubsection{Question and Answer Generation.}
% For each video, we segment it according to the action-level annotations provided by MGIT, ensuring segment boundaries align with genuine causal transitions in the narrative. This alignment guarantees that each segment reflects a distinct causal event or state change, providing a solid foundation for stepwise reasoning. Detailed segment and its descriptions serve as input for GPT-4o \cite{gpt4o}. GPT-4o is then prompted to generate candidate QA pairs for both descriptive understanding and explicit stepwise causal reasoning tasks. The prompts are carefully designed to encourage diversity, clarity, and alignment with the reasoning tasks. Following automatic generation, human annotators meticulously review all candidate QA pairs, filtering out ambiguous, irrelevant, or low-quality items. Annotators further refine the wording, ensure factual accuracy, and verify that each question is appropriately grounded in its corresponding video segment and, for causal reasoning tasks, in the correct causal chain. This two-stage process leverages the efficiency of LLMs while maintaining high annotation quality and task validity.
For each video, we segment it based on MGIT's action-level annotations, ensuring boundaries align with genuine causal transitions. This guarantees each segment reflects a distinct causal event or state change, foundational for stepwise reasoning. Detailed segment descriptions are input to GPT-4o \cite{gpt4o}, which generates candidate QA pairs for both descriptive understanding and explicit stepwise causal reasoning. Prompts are designed for diversity, clarity, and task alignment. Human annotators then meticulously review and refine all candidate QA pairs, filtering ambiguous or low-quality items, ensuring factual accuracy and proper grounding in the video segments and causal chains. This two-stage process leverages LLM efficiency while maintaining high annotation quality and task validity.

\subsubsection{Taxonomy-Based Distractor Generation.}
% For each multiple-choice question, the correct answer is taken from our refined QA pair. To address inconsistencies in difficulty and quality, and to introduce explicit “error type” design, we propose a novel taxonomy-based distractor generation approach. For each question, we first define several typical error types (e.g., temporal confusion, causal misattribution, object misrecognition, irrelevant but plausible events). Distractor options are then systematically generated to cover these categories. GPT-4o \cite{gpt4o} is instructed to propose plausible but incorrect alternatives that are contextually relevant and semantically similar to the correct answer, specifically aligned with the defined error types. Human annotators then review and edit these distractors, ensuring they are non-trivial, factually sound, and that each distractor fits its intended error type and maintains comparable plausibility. During evaluation, the order of options was randomized for each question to mitigate positional bias. This process maximizes the challenge, prevents models from selecting answers based on superficial cues, and makes the distractors challenging, diverse, and diagnostically informative for reasoning evaluation, further enhancing the benchmark's rigor and reliability. We provide an example of the QA pairs in the CausalStep benchmark in Figure \ref{fig:case}. For detailed information about the error types, please refer to Appendix A.

\begin{figure}[htbp!]
  \centering   
  \includegraphics[width=\linewidth]{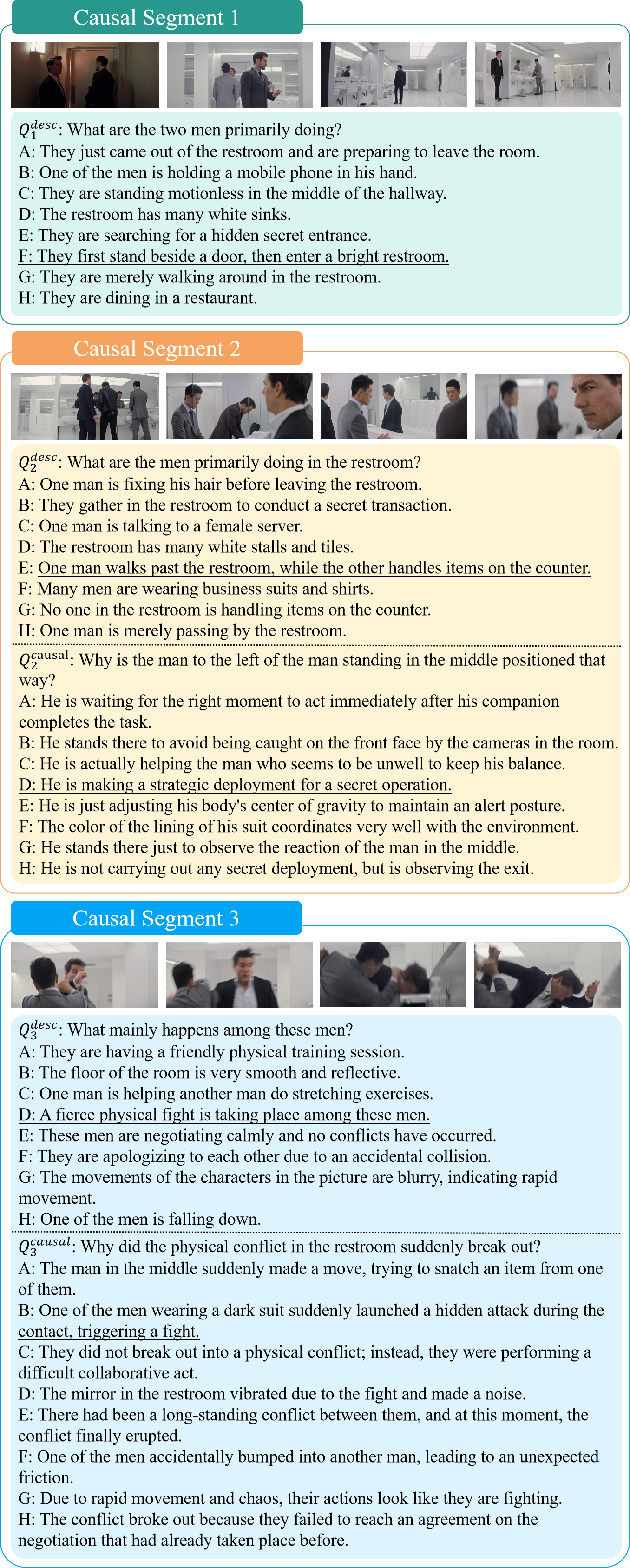}
  \caption{Example of the QA pairs in the CausalStep benchmark. The figure shows three consecutive causal segments, along with their corresponding Descriptive Understanding $Q^{desc}_i$ and Stepwise Causal Reasoning $Q^{causal}_i$.}
  \label{fig:case}
\end{figure}

For each multiple-choice question, the correct answer comes from our refined QA pair. To ensure consistent difficulty and introduce explicit “error type” design, we propose a novel taxonomy-based distractor generation approach. We first define typical error types (e.g., temporal confusion, causal misattribution, object misrecognition). Distractor options are then systematically generated by GPT-4o \cite{gpt4o} to be plausible, incorrect, contextually relevant, and semantically similar alternatives, specifically aligned with these error types. Human annotators meticulously review and edit these distractors, ensuring they are non-trivial, factually sound, and maintain comparable plausibility while fitting their intended error type. Option order is randomized during evaluation to mitigate positional bias. This process maximizes challenge, prevents models from relying on superficial cues, and makes distractors diverse and diagnostically informative, enhancing the benchmark's rigor. Figure \ref{fig:case} shows a QA pair example; Appendix A details the error types.

\begin{table}[htbp]
 \centering
 \vspace{-5pt}
 \caption{Statistics of the CausalStep benchmark.}
 \vspace{-5pt}
 \begin{tabular}{lc}
  \toprule
  \textbf{Statistic} & \textbf{Value} \\
  \midrule
  \#Videos & 100 \\
  Video duration (mean) & 430.5 s \\
  Video duration (min / max) & 149 s / 994.4 s \\
  \#QA pairs & 1,852 \\
  QA type & Multiple-choice \\
  Options per question & 8 \\
  \#Categories & 6 \\
  Avg. segments per video & 8.76 \\
  Segments per video (min / max) & 2 / 51 \\ % Clarified unit
  Annotation & AI-assisted + Manual \\
  Distractor design & Error-type taxonomy \\
  Descriptive QA pairs & 926 \\
  Reasoning QA pairs & 926 \\
  \bottomrule
 \end{tabular}
 \label{tab:causalstep_stats}
 \vspace{-10pt}
\end{table}

\label{sec:expr}
\begin{table*}[t!]
 \centering
 \vspace{-5pt}
 \caption{Performance comparison of open-source and proprietary models on CausalStep using seven diagnostic metrics. CSR (\%): chain success rate (↑), AMCL: average maximum chain length per video (↑), MCL: global maximum chain length (↑), RF: average restart frequency (↓), WS: weighted score (↑), DUA (\%): descriptive understanding accuracy (↑), and ICRA (\%): isolated causal reasoning accuracy (↑). The last three rows report the best model performance, average human performance, and theoretical maximum (upper bound for perfect chains).}
 \vspace{-5pt}
 \begin{tabular*}{\textwidth}{@{\extracolsep{\fill}}lccccccc}
  \toprule
  \textbf{Model} & \textbf{CSR(\%) $\uparrow$} & \textbf{AMCL $\uparrow$} & \textbf{MCL $\uparrow$} & \textbf{RF $\downarrow$} & \textbf{WS $\uparrow$} & \textbf{DUA(\%) $\uparrow$} & \textbf{ICRA(\%) $\uparrow$} \\
  \midrule
  \multicolumn{8}{c}{\emph{Open-source models}} \\
  \midrule
  LLaVA-Onevision & 7 & 5.20 & 4 & 3.14 & 30.85 & 67.1 & 15.2 \\
  Video-LLaVA & 10 & 5.15 & 5 & 3.13 & 32.94 & 68.6 & 20.1 \\
  Phi4-multimodal-instruct & 13 & 5.33 & 4 & 3.01 & 33.78 & 70.1 & 21.4 \\
  Qwen2.5-VL-7B & 16 & 5.61 & 9 & 2.68 & 35.42 & 71.0 & 21.8 \\
  InternVL3-8B & 19 & 5.59 & 8 & 2.87 & 35.26 & 69.2 & 23.1 \\
  Gemma3-12b-it & 21 & 5.53 & 11 & 2.81 & 36.22 & 72.9 & 24.5 \\
  InternVL3-38B & 24 & 5.75 & 13 & 2.57 & 36.89 & 75.3 & 25.1 \\
  Qwen2.5-VL-72B & 26 & 5.89 & 17 & 2.47 & 37.69 & 76.1 & 25.2 \\
  Gemma3-27b-it & 29 & 5.94 & 20 & 2.42 & 37.64 & 77.7 & 26.3 \\
  \midrule
  \multicolumn{8}{c}{\emph{Proprietary models}} \\
  \midrule
  Gemini-2.0-Flash & 31 & 6.04 & 21 & 2.45 & 39.60 & 79.4 & 27.1 \\
  Claude-3.5-Sonnet-20241022 & 35 & 5.87 & 23 & 2.37 & 38.58 & 80.9 & 28.5 \\
  GPT-4o-2024-11-20 & 39 & 5.94 & 23 & 2.17 & 38.88 & 80.0 & 29.7 \\
  Gemini-2.0-Flash-thinking & 41 & 6.15 & 25 & 2.15 & 40.65 & 81.1 & 30.2 \\
  GPT-4.1-2025-04-14 & 42 & 6.63 & 26 & 1.85 & 45.59 & 82.8 & 32.3 \\
  Gemini-2.5-Flash & 48 & 6.90 & 27 & \textbf{1.68} & 47.63 & 84.6 & 36.2 \\
  o4-mini-2025-04-16 & \textbf{51} & \textbf{7.19} & \textbf{30} & 1.69 & \textbf{55.06} & \textbf{85.2} & \textbf{39.8} \\
  \midrule
  \emph{Best Performance of Models} & 51 & 7.19 & 30 & 1.68 & 55.06 & 85.2 & 39.8 \\
  \emph{Human} & 79 & 8.03 & 46 & 0.74 & 62.39 & 92.0 & 76.8 \\
  \emph{Maximum} & 100 & 8.76 & 51 & 0 & 68.76 & 100.0 & 100.0 \\
  \bottomrule
 \end{tabular*}
 \label{tab:benchmark_result}
 \vspace{-10pt}
\end{table*}

\subsection{Benchmark Statistics}
CausalStep is a comprehensive benchmark comprising 100 videos (average duration 430.5 seconds, ranging from 149 to 994.4 seconds) across 6 diverse categories. Each video is meticulously segmented into an average of 8.76 causal segments (ranging from 2 to 51 segments per video), forming the basis for our stepwise reasoning tasks. The benchmark features a total of 1,852 multiple-choice QA pairs, evenly split between 926 descriptive understanding questions and 926 causal reasoning questions. Each question averages 8 options, including 1 correct answer and 7 challenging distractors meticulously designed using a novel error-type taxonomy. The entire annotation process employs a hybrid AI-assisted and manual review approach to ensure high data quality. Key statistics are summarized in Table \ref{tab:causalstep_stats}, providing a detailed overview of the benchmark's scale and characteristics. For more statistical information about the CausalStep benchmark, please refer to Appendix D.

\section{Experiments}

\subsection{Settings}
\subsubsection{Baselines.}
We evaluate a comprehensive set of models, including both leading proprietary and open-source MLLMs. Specifically, the closed-source baselines include the GPT series (GPT-4o~\cite{gpt4o}, GPT-4.1-2025-04-14~\cite{gpt41} and o4-mini-2025-04-16~\cite{o4mini}), the Gemini series (Gemini-2.0-Flash, Gemini-2.0-Flash-thinking~\cite{gemini}, Gemini-2.5-Flash~\cite{gemini2.5}), Claude-3.5-Sonnet-20241022~\cite{claude35}. The open-source baselines include the Qwen series (Qwen2.5-VL 7B/72B-Instruct~\cite{qwen25}), the Gemma series (Gemma3 12B/27B~\cite{gemma3}), the InternVL series (InternVL3 8B/38B~\cite{internvl3}), the LLaVA series (LLaVA-OneVision-7B~\cite{llavaov}, Video-LLaVA-7B~\cite{videollava}), and Phi4-multimodal-Instruct~\cite{phi4}.

For a fair comparison, all models use consistent video frame sampling strategies and input formats, including the same number of frames. Implementation details and prompts are in Appendix E. We also conduct human experiments to establish an upper bound and analyze the human-MLLM reasoning gap. More details please refer to Appendix F.

\subsubsection{Metrics.}
To comprehensively evaluate model performance on explicit stepwise causal reasoning, we employ five key metrics and two supplementary indicators:

\begin{itemize}
\item \textbf{Chain Success Rate (CSR)}: Proportion of videos where the model completes the entire reasoning chain without errors, reflecting global consistency and long-range reasoning ability.
\item \textbf{Average Maximum Chain Length (AMCL)}: Average length of the longest uninterrupted reasoning chains achieved across all videos, indicating typical reasoning depth.
\item \textbf{Maximum Chain Length (MCL)}: The single longest uninterrupted reasoning chain achieved in any video, denoting peak reasoning depth.
\item \textbf{Restart Frequency (RF)}: How often reasoning chains are interrupted and restarted due to incorrect answers; lower RF indicates greater robustness.
\item \textbf{Weighted Score (WS)}: Scores increase for later steps in a correct chain (as detailed in task, rewarding longer, sustained reasoning sequences.
\end{itemize}
Two supplementary indicators further dissect model capabilities:
\begin{itemize}
\item \textbf{Descriptive Understanding Accuracy (DUA)}: Accuracy on isolated descriptive questions for each segment, measuring foundational visual perception.
\item \textbf{Isolated Causal Reasoning Accuracy (ICRA)}: Accuracy on causal questions when only the current segment is provided, revealing reliance on local evidence for causal reasoning.
\end{itemize}
% Together, these seven metrics offer a fine-grained, multi-faceted assessment of both causal reasoning proficiency and foundational content understanding.

\subsection{Main Results}
Table~\ref{tab:benchmark_result} summarizes the performance of a diverse set of open-source and proprietary models, alongside human baselines, on CausalStep using seven diagnostic metrics. Overall, we observe a clear stratification of model capabilities. Proprietary models consistently outperform open-source models across all metrics, with o4-mini-2025-04-16 achieving the best performance among all evaluated models. Specifically, o4-mini attains a Chain Success Rate (CSR) of 51\%, an Average Maximum Chain Length (AMCL) of 7.19, a Maximum Chain Length (MCL) of 30, and a Weighted Score (WS) of 55.06, while maintaining a relatively low Restart Frequency (RF) of 1.69. In contrast, the top open-source model, Gemma-3-27b-it, lags significantly with a CSR of 29\%, AMCL of 5.94, and MCL of 20.

Human participants set a strong performance ceiling, achieving a CSR of 79\%, AMCL of 8.03, MCL of 46, the highest WS, and the lowest RF among all evaluated entities. For the single-step metrics, Descriptive Understanding Accuracy (DUA) and Isolated Causal Reasoning Accuracy (ICRA), proprietary models again show superior performance over open-source counterparts. However, all models demonstrate a substantial performance gap compared to human participants, particularly evident in ICRA, highlighting the inherent challenge of causal reasoning without the benefit of full contextual understanding derived from a successful chain.

\subsection{Analysis}
\subsubsection{MLLMs' Strengths and Limitations in CausalStep.}
Our findings reveal a persistent gap between current MLLMs and human-level performance across all diagnostic metrics, underscoring the demanding nature of the CausalStep benchmark. While the best-performing model, o4-mini, achieves a CSR of 51\% and a MCL of 30, human participants reach significantly higher levels with 79\% CSR and an MCL of 46. This disparity is further emphasized by the AMCL and WS, where human performance consistently exceeds that of all models. These results collectively indicate that existing models largely struggle to maintain long, uninterrupted reasoning chains, exhibiting a propensity for frequent interruptions as evidenced by their comparatively higher RF values. This suggests a primary limitation in sustaining deep, multi-step causal reasoning over extended video sequences.

\subsubsection{Open-source vs. Proprietary Models.}
A clear stratification in capabilities is observed between open-source and proprietary models. Proprietary models, exemplified by o4-mini and the Gemini series, consistently demonstrate superior performance across all metrics. For instance, o4-mini surpasses the best open-source model, Gemma-3-27b-it, by a notable 22 points in CSR and 10 points in MCL. This divergence highlights the impact of more extensive resources, potentially larger and more diverse training data, and sophisticated architectural designs prevalent in proprietary systems. However, despite their lead, even the most advanced proprietary models remain considerably behind human-level performance, signaling that significant advancements are still required to bridge this fundamental reasoning gap.

\subsubsection{Single-step Understanding vs. Stepwise Reasoning.}
A granular examination of the single-step metrics—DUA and ICRA—sheds light on distinct challenges. While top models achieve relatively high DUA (up to 85.2\%), indicating competence in isolated perceptual understanding, their ICRA remains markedly lower (best model at 39.8\%). This stark contrast, coupled with humans' strong ICRA (76.8\%), underscores a critical limitation: current models struggle to perform accurate causal reasoning when presented solely with an isolated segment pair, without the benefit of a preceding, correctly established reasoning chain. This discrepancy validates the necessity of CausalStep's stepwise protocol, which inherently evaluates the ability to build and leverage contextual reasoning through sequential reasoning.

\subsubsection{Implications for Advancing Video Reasoning.}
Collectively, these findings firmly establish CausalStep as a rigorous and discriminative benchmark, effectively diagnosing the strengths and weaknesses of MLLMs in causal video reasoning. The persistent human-model gap, especially in long-range and stepwise causal reasoning, highlights critical future research avenues: enhancing model memory for context integration, developing robust and explicit causal reasoning mechanisms, and designing training curricula emphasizing multi-step, context-dependent, and error-recovering reasoning. We believe CausalStep will inspire advancements toward human-level causal intelligence in complex videos.

\section{Conclusion}
\label{sec:conclusion}
In this work, we introduce CausalStep, a diagnostic benchmark for explicit stepwise causal reasoning in videos. By segmenting videos into causally linked units and enforcing a strict stepwise QA protocol, CausalStep enables rigorous evaluation of sequential causal reasoning. Our experiments show a huge gap between current MLLMs and human-level performance, particularly in maintaining long reasoning chains and integrating causal context. While proprietary models outperform open-source ones, neither approaches human ability on the challenging task. These findings underscore the value of CausalStep for diagnosing model limitations and highlight the need for more robust and causally aware video reasoning systems.

\clearpage
\bibliography{aaai2026}
\end{document}